\documentclass[a4paper]{./styles/svproc}
\usepackage{url}
\usepackage{graphicx,subfigure}
\usepackage{breakurl}
\usepackage{cite}
\usepackage{float}
\usepackage{amsmath} 
\usepackage[linesnumbered,ruled,vlined]{algorithm2e}
\usepackage{todonotes}
\usepackage{etoolbox}
\usepackage{multirow}

\begin{document}

\mainmatter              
\title{\vspace{-1em}{\normalsize 2024 IEEE International Symposium of Robotics Research (ISRR)\\[1.5em]}An Energy-Aware Routing Algorithm for \\
Mobile Ground-to-Air Charging}
\titlerunning{Mobile Charging}  
%
\author{Bill Cai, Fei Lu, Lifeng Zhou\thanks{Corresponding author}}
\authorrunning{Bill Cai, et al.} 
%
%
\institute{Drexel University, Philadelphia, PA 19104, USA\\
\email{\{sc3568,fl345,lz457\}@drexel.edu}\\ }

\maketitle              
\vspace{-7mm}
\begin{abstract}
We investigate the problem of energy-constrained planning for a cooperative system consisting of an Unmanned Ground Vehicle (UGV) and an Unmanned Aerial Vehicle (UAV). In scenarios where the UGV serves as a mobile base to ferry the UAV and as a charging station to recharge the UAV, we formulate a novel energy-constrained routing problem. To tackle this problem, we design an energy-aware routing algorithm, aiming to minimize the overall mission duration under the energy limitations of both vehicles. The algorithm first solves a Traveling Salesman Problem (TSP) to generate a guided tour. Then, it employs the Monte-Carlo Tree Search (MCTS) algorithm to refine the tour and generate paths for the two vehicles, taking into account multiple physical constraints such as charging speed, total energy expenditure, travel time, and other operational requirements. We evaluate the performance of our algorithm through extensive simulations and a proof-of-concept experiment. The results show that our algorithm consistently achieves near-optimal mission time and maintains fast running time across a wide range of problem instances. 
\end{abstract}
\vspace{-8 mm}
\section{Introduction}
\begin{figure}[h]
\centering{
\subfigure[]{\includegraphics[width=0.4\columnwidth]{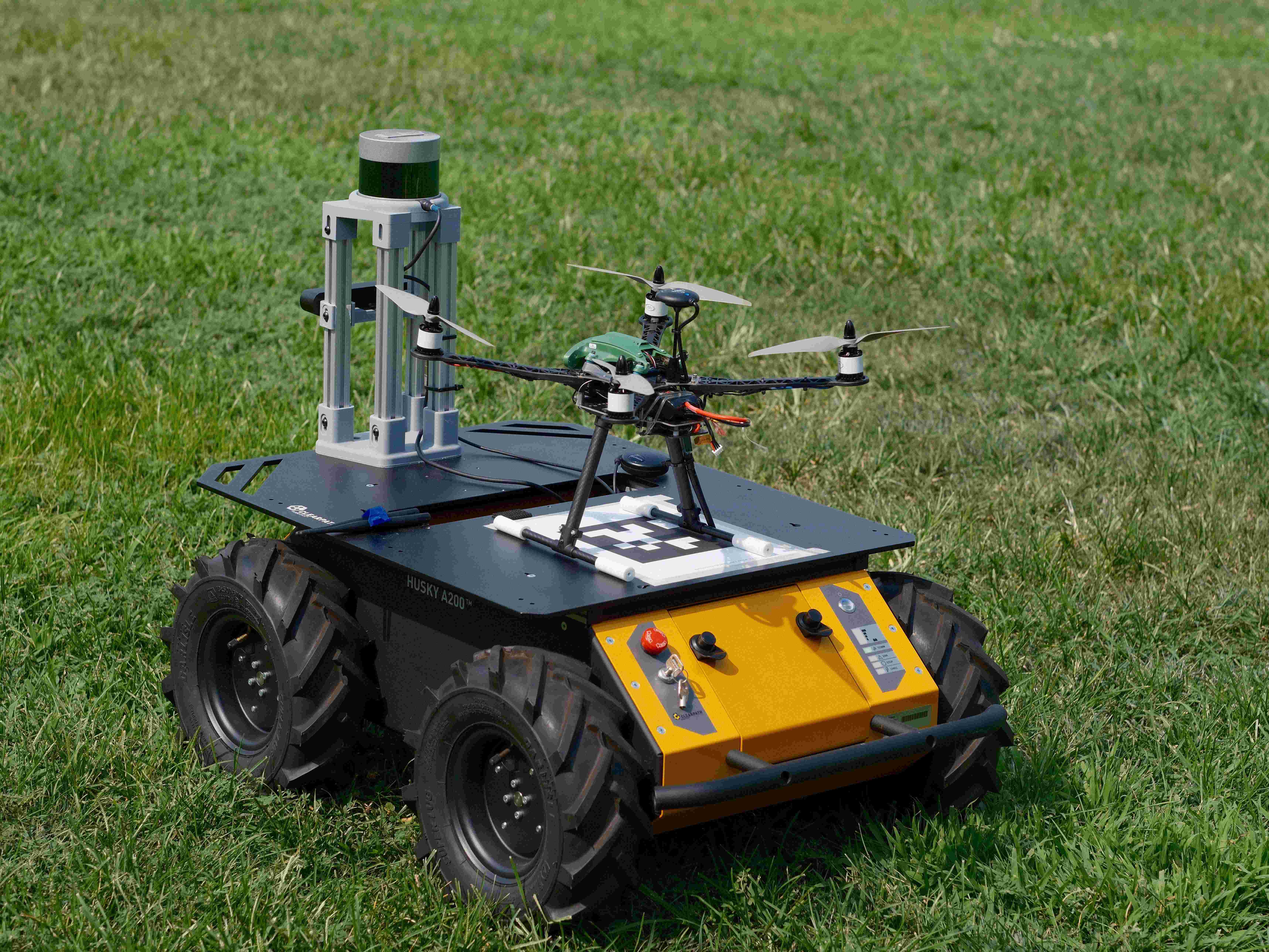}}
\subfigure[]{\includegraphics[width=0.4\columnwidth]{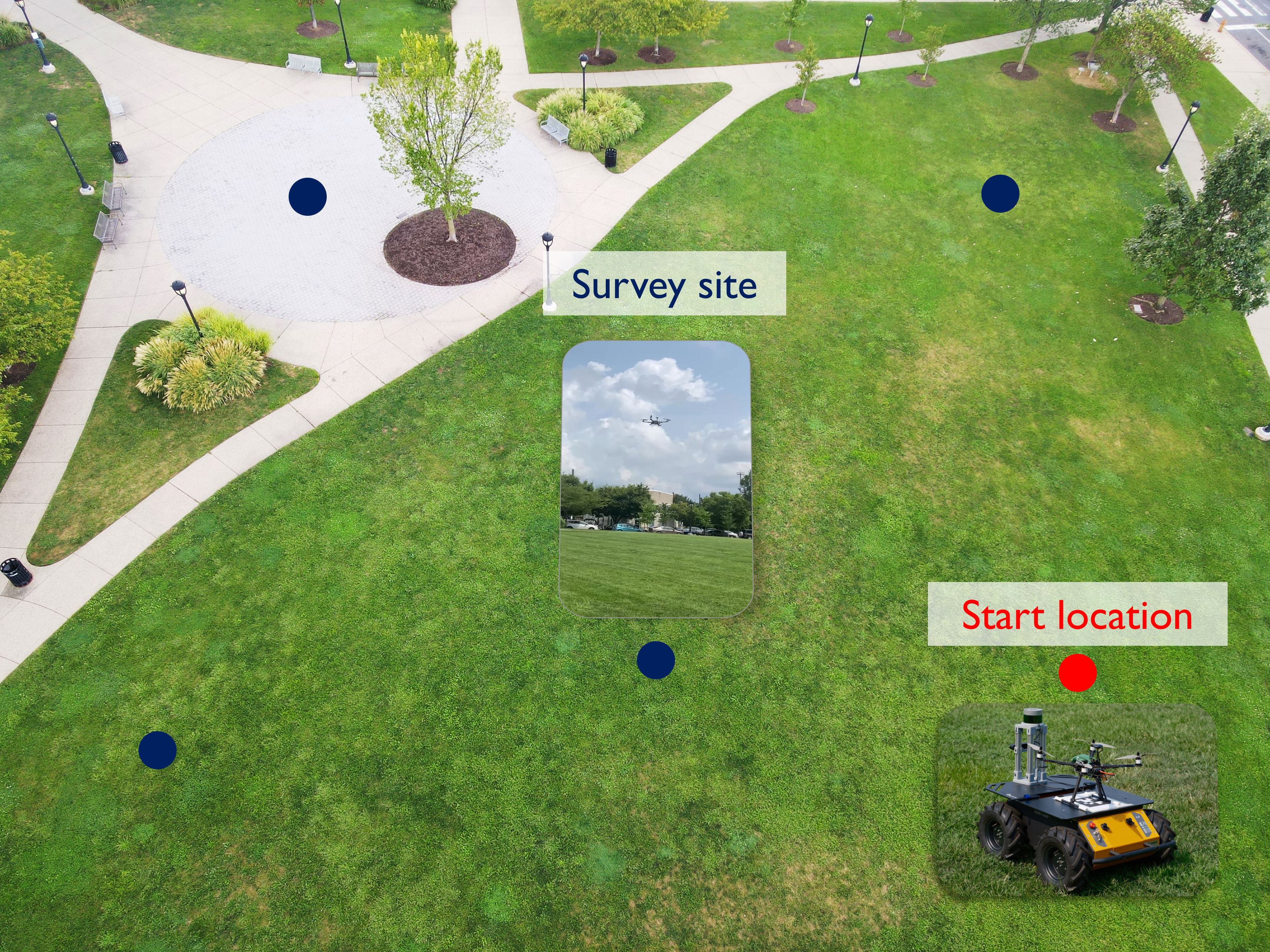}}
}
\label{fig:drexlpark}
\caption{A cooperative autonomous platform (a) consisting of a UGV (Clearpath Husky rover) and a UAV (ModalAI Sentinel drone) is tasked to survey five sites (blue dots) scattered on the Drexel Park (b).}
\end{figure}

\vspace{-1 mm}
Unmanned Aerial Vehicles (UAVs) have become indispensable across a plethora of applications, from surveillance \cite{kingston2008decentralized} and package delivery \cite{Package} to infrastructure inspection\cite{burri2012aerial}, environmental monitoring\cite{corrales2012volcano}, and precision agriculture\cite{ribeiro2021multi}. Their potential is especially significant for the continuous monitoring of dynamic environments, such as air quality sampling, border security, and the visual inspections of power plants and pipelines.  In these persistent surveillance tasks, UAVs are constantly navigating the environment to ensure high-quality and frequent sensing in specified regions. For example, in this paper, we consider an energy-constrained information collection scenario where a UAV is tasked to sequentially survey multiple sites of interest (Fig.~\ref{fig:drexlpark}). For continuous operation, the UAVs' limited battery life poses a significant challenge, necessitating the development of optimized planning strategies to extend mission duration.  To overcome this challenge, a substantial body of research has emerged. Solutions proposed include automated battery swapping mechanisms \cite{swapping}, energy-efficient low-level controllers \cite{palossi2016energy}, tactical low-level path planning \cite{chodnicki2022energy}, and charging through Unmanned Ground Vehicles (UGVs) \cite{bin2017autonomous}. 


Alongside, there has been a distinct emphasis on high-level path planning \cite{mathew2015multirobot,mobilerecharg}, with a focus on energy optimization, especially when integrating UAVs with UGV charging systems. One notable work introduces the concept of employing dedicated charging robots for recharging UAVs during their long-term missions \cite{mathew2015multirobot}. Their approach is to discretize the UAV's 3D flight trajectories, projecting them as charging points on the ground. This representation enables an abstraction onto a partitioned graph, from which paths for charging robots are derived. Their methods exploit both integer linear programs and a transformation to the Traveling Salesman Problem (TSP) to craft rendezvous planning strategies. Another significant contribution to this domain is the exploration of energy-constrained UAV missions, with a focus on minimizing tour time \cite{mobilerecharg}. The authors propose a unique TSP variant that incorporates mobile recharging stations. Their algorithm determines not only the sequence of site visits but also the optimal timings and locations for UAVs to dock on charging stations. They examine multiple charging scenarios, including stationary and mobile charging stations, delivering practical solutions using Generalized TSP. A more generalized view of persistent surveillance using energy-limited UAVs, supported by UGVs as mobile charging platforms, is presented in \cite{lin2022robust}. Recognizing the inherent NP-hard nature \cite{2008routing} of this combinatorial optimization problem, the authors propose a robust approximate algorithm. Their strategy revolves around forming uniform UAV-UGV teams, partitioning the surveillance environment based on UAV fuel cycles, and maintaining cyclic paths that UAVs and UGVs traverse. These works highlight the complexities and challenges in planning UAV missions with recharge constraints. 


However, these studies often overlook the energy constraints of the UGVs themselves, a factor that must be considered in real-world applications. In this paper, we address these overlooked aspects by
adopting a comprehensive approach that takes into account the energy requirements of \textit{both} UAVs and UGVs. Specifically, we introduce an energy-aware routing algorithm that combines TSP solutions with the Monte-Carlo Tree Search (MCTS) algorithm for optimal mission planning.

\noindent \textbf{Contributions} In this paper, we design and evaluate a novel strategy addressing the intertwined energy and routing challenges in a heterogeneous UAV-UGV system. Our main contributions are as follows.

\begin{itemize}
    \item \textit{Problem formulation:} We formulate an energy-constrained routing problem for a cooperative system composed of a UGV and a UAV to survey multiple sites in minimal time and under energy constraints for both vehicles. We are the first to address energy constraints of \textit{both} UAV and UGV in this cooperative routing problem.
    

    \item \textit{Approach:}
    Our algorithm begins by using a TSP solver to create an initial route. It then employs the MCTS to explore optimal paths for both the UGV and UAV. This process aims to reduce the total mission duration while adhering to energy limitations. Unlike previous methods, our approach comprehensively accounts for real-world factors including the UGV's energy consumption rate when operating alone versus with the UAV on it, the UAV's charging time, and real-time energy levels of both vehicles throughout the mission.
    
    \item \textit{Results:}
    We conduct both extensive situations and a proof-of-concept experiment to demonstrate the effectiveness and efficiency of our algorithm. The results show that our algorithm generates near-optimal paths for the UGV and UAV and maintains fast running speed even when the number of sites is $50$.

\end{itemize}
\vspace{-5 mm}
\section{Problem formulation}\label{sec:problem}
\vspace{-1 mm}
Considering an environment with $N$ distinct target sites, we aim to devise a routing strategy for a UAV and UGV pair to survey these sites efficiently.  We seek high-level routing algorithms to minimize the total mission time  $T_{\text{total}}(\mathbf{r})$ for sequentially surveying all these sites, and ensuring both vehicles return to their starting point within the energy constraints. We outline relevant notations in Table 1, followed by a detailed discussion on the mission objective, operational dynamics of UGV and UAV, and energy constraints. We then formally present our energy-constrained routing problem.

\begin{problem}
\vspace{-3mm}
   \begin{align} 
\begin{split}
    & \min_{} \ \  T_{\text{total}}  \\ 
    \text{s.t.}~ \
    & e_{\text{g},t} > 0, \ \ \\
    & e_{\text{a},t} > 0, \ \ \\
    & e_{\text{g},T} \geq 0, \ \ \\
    & e_{\text{a},T} \geq 0. \ \
\end{split}
\end{align}
\end{problem}
The first two constraints ensure that both the energy of UGV and UAV should be above zero during the mission. The last two constraints tell that the two vehicles can use up their energy when returning to the start location. 

\begin{table}[H]
\centering
\begin{tabular}{|c|l|}
\hline
\textbf{Symbol} & \textbf{Definition} \\
\hline
$E_{\text{Gmax}}$ & Maximum energy (i.e., battery capacity in mAh) of UGV. \\
$E_{\text{Amax}}$ & Maximum energy (i.e., battery capacity in mAh) of UAV. \\
$e_{\text{g},t}$ & The energy of UGV at time $t$. \\
$e_{\text{a},t}$ & The energy of UAV at time $t$. \\
$C_{\text{a}}$ & Flying cost of UAV mAh per kilometer. \\
$C_{\text{s}}$ & Surveying cost of UAV mAh per hours. \\
$C_{\text{g}}$ & Moving cost of UGV mAh per kilometer. \\
$C_{\text{ga}}$ & Moving cost of UGV mAh per kilometer when ferrying UAV. \\
$R_{\text{c}}$ & Charging speed in mAh per kilometer. \\
$V_{\text{a}}$ & Speed of UAV in kilometer per hour. \\
$V_{\text{g}}$ & Speed of UGV in kilometer per hour. \\
$T_{\text{survey}}$ & Time taken for surveying a site in hours. \\
$T_{\text{Gwait}}$ & Time taken by UGV waiting for UAV for all sites in hours. \\
$T_{\text{total}}$ & Total mission time in hours.\\
\( r \) & The radius of the circle in kilometer. \\
\( O \) & The center of the circle in kilometer. \\
\( AB \) & The chord that UGV traverses in kilometer. \\
\( S \) & The distance from \( A \) where the UGV starts in kilometer. \\
\( P \) & The point where the UAV and UGV rendezvous. \\
\( M \) & The midpoint of the chord \( AB \). \\
\hline
\end{tabular}
\vspace{1mm}
\caption{Notations used in the paper.}
\end{table}
\vspace{-15 mm}
\subsection{Mission objective}
\begin{figure}[h]
\centering
\subfigure{\includegraphics[width=0.4\columnwidth]{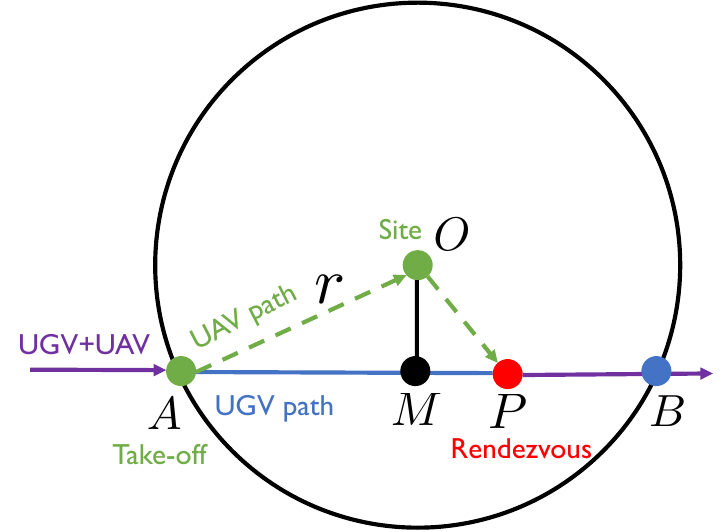}}
\caption{\small An illustration of the calculation for the survey radius \(r\) and the rendezvous point \(P\) for the cooperative operation of UAV and UGV at a site. The circle, centered at site \(O\) with radius \(r\), represents the UAV's operational area based on its allocated energy for the site. Points \(A\) and \(B\) mark the chord's endpoints across which the UGV moves, with \(M\) denoting the midpoint. The UAV departs from \(A\) and lands on the UGV at point \(P\), following the \(A\to O \to P\) flight path. The line \(MO\) serves as a perpendicular bisector to chord \(AB\).}

\label{fig:Tri} 
\vspace{-5 mm}
\end{figure}

\vspace{-1 mm}
\label{subsec:mission}
The mission asks for a cooperative system of a UGV and a UAV to survey multiple sites of interest. The UGV serves a dual role---as a mobile base to ferry the UAV and as a mobile charging station to charge the UAV. As the UGV approaches a site, the UAV commences its take-off and heads to surveying the site. Once it completes the survey, it returns and lands on the UGV. This represents one phase of cooperative operation, corresponding to the survey of one site (Figure~\ref{fig:Tri}).\footnote{In this paper, we omit the energy and time required for take-off and landing. However, these factors can be easily incorporated into the problem and the proposed solution~(Section~\ref{sec:approach}).}

Then the next phase starts and the UGV mules the UAV to survey the next site. The same operation continues until all sites are surveyed and the two vehicles return to the start location. 

In particular, upon the UAV finishing surveying a site, the two vehicles identify a rendezvous point along the UGV's path for the UAV to land. We allow the UGV to arrive earlier at the rendezvous point to wait for the UAV instead of the opposite. That is because, if the UAV arrives earlier and hovers around, it may drain its energy and not be able to land on the UGV, which results in an entire mission failure.

Therefore, the total mission time $T_{\text{total}}$ can be defined as the summation of the UGV's travel time and the total waiting time it waits for the UAV to return to it after surveying each site. $T_{\text{total}} = \frac{d_{\text{UGV}}}{V_{\text{g}}}  + T_{\text{Gwait}}$ where the UGV's travel time is computed as $d_{\text{UGV}}/V_{\text{g}}$  where \(d_{\text{UGV}}\) represents the distance traveled by the UGV, and \(V_{\text{g}}\) is the UGV's speed. Our primary goal is to minimize the total mission time.  In other words, we aim to optimize the efficiency of the UAV-UGV cooperative mission and ensure the timely completion of the survey task.
\vspace{-2 mm}
\subsection{Cooperative operation for site survey}\label{subsec:coop_opera}
\vspace{-1 mm}
We introduce in detail a phase of cooperative operation of UGV and UAV. Here, we consider the UAV to have a set of discrete energy levels based on its current battery level, which follows the same setting for energy management in the previous work~\cite{lin2022robust}. For example, if the UAV's current battery level is 100\%, the set of energy levels it has can be defined as $[20\%, 40\%, 60\%, 80\%, 100\%]$. 
Our approach optimizes energy use by selecting the best allocation from a set of candidate energy levels, ensuring it meets the UAV's energy availability, adheres to the mission's overall energy limits, and minimizes total mission time.

To utilize this, our approach involves converting the chosen energy level for the UAV into a range. This range should be sufficient for the UAV to make a round trip from the UGV to the site, as well as to conduct a survey of the site. By doing this conversion, we ensure that after reaching the site and completing its survey, the UAV retains enough energy to land anywhere within a circle, with the site at the center and the computed range as its radius, as shown in Figure~\ref{fig:Tri}. The range of the UAV can be determined by subtracting the energy used for the survey from the UAV's available energy, and then dividing by its per kilometer flying cost $C_{\text{a}}$.

Given an energy allocation of the UAV, $e_{\text{a}} \in \{10\%\cdot E_{\text{Amax}}, 20\%\cdot E_{\text{Amax}}, \ldots, 100\%\cdot E_{\text{Amax}}\}$, the radius can be calculated by

\begin{equation}
r = \frac{{e_{\text{a}} - C_{\text{s}}T_{\text{survey}}}}{2C_{\text{a}}}.
\label{eq:maxAdistance}
\end{equation}

Notably, each energy allocation corresponds to the radius of a circle where the UAV can complete a round trip between any point on the circle and the circle's center plus the survey at the center by using the allocated energy. 

Figure~\ref{fig:Tri} shows one phase of cooperative operation between the UGV and the UAV. The UAV takes off from the UGV, flies to the site for survey, and returns to the UGV at the rendezvous point (red dot). Meanwhile, the UGV moves from the take-off point to the rendezvous point to meet the UAV, and then ferries it to the next take-off point for the next phase. 

\vspace{-5 mm}
\subsection{Energy constraint}
\vspace{-1 mm}
\label{subsec:energycon}
We consider the energy constraint for both the UGV and UAV. The UGV is required to not run out of power before returning to the start location. Notably, the UGV consumes more energy when transporting the UAV, even without charging, with the energy cost rate denoted by \(C_\text{ga}\). This rate is used to compute the UGV's energy level during the tour.

The energy allocated for UAV recharging is determined by three factors: the UGV's available energy \(e_g\), the UAV's energy deficiency, and the remaining distance to the next waypoint. These factors affect the maximum potential charge the UAV can receive, calculated by the charging speed \(R_c\) over distance. The smallest value among these three factors will dictate the actual energy allocated for charging. We assume a recharging efficiency of 100\%. This can be easily adjusted according to application contexts by employing a corresponding battery-charging model. 

Charging is presumed to initiate immediately upon the UAV's landing on the UGV. The amount of charge is determined as the minimum among the potential charge, the UAV's energy deficiency, and the UGV's energy capacity. This is formally defined by ${\min}\{R_{\text{c}} d_{\text{UAV+UGV}}, E_{\text{Amax}} - e_{\text{a},t}, e_{\text{g},t}\}$.   Notably, $e_{\text{g},t}$ becomes the limiting factor, the energy required for the UGV to return to the start location and the additional energy cost of carrying the UAV to the next possible release point must be subtracted from its available energy to determine the actual amount of energy that can be allocated for charging the UAV.
While for UAV, since it can be recharged by the UGV, it must be able to return to the UGV for recharging before using up its energy during the entire mission. The violation of either UGV's or UAV's energy requirement indicates a mission failure.

In summary, the main problem considered in this paper is to compute the optimal routes for the UGV and UAV to minimize the total mission time (Section~\ref{subsec:mission}) while satisfying all energy constraints (Section~\ref{subsec:energycon}). Notably, 
finding the optimal solutions requires evaluating all possible tours of surveying the sites and all possible combinations of the battery levels across the sites.  Clearly, the problem has combinatorial complexity and is NP-hard. To tackle the problem, we design an energy-aware routing algorithm that leverages TSP solution for computing a tour and MCTS for battery level selection in Section~\ref{sec:approach}. 



\vspace{-5 mm}
\section{Approach} \label{sec:approach}
\vspace{-2 mm}
As discussed in Section~\ref{sec:problem}, the problem of planning paths for the UGV and UAV with energy constraints is NP-hard. Solving the problem directly is intractable. Therefore, we decompose the problem into two smaller subproblems. Specifically, the first subproblem asks for computing a tour for the two vehicles to efficiently visit and survey all the sites. Subsequently, in the second problem, we focus on allocating the optimal energy level for the UAV at each site to minimize the total mission time with the energy constraints considered. We develop an energy-aware UGV-UAV routing algorithm that solves these two subproblems in two steps, as shown in Algorithm~\ref{alg:routing}. Next, we introduce Algorithm~\ref{alg:routing}'s two steps in more detail. 
\vspace{-5 mm}
\subsection{The first step of Algorithm~\ref{alg:routing}: TSP}
\vspace{-1 mm}
The first step of Algorithm~\ref{alg:routing} (Line~\ref{line:tsp}) aims to solve the first subproblem that asks for computing a survey tour of all the sites. Given the overall objective is to minimize the total mission time, it is reasonable to seek the shortest tour for the two vehicles. Then, this subproblem can be framed as a TSP  that entails finding the shortest possible route that visits a list of sites precisely once before returning to the origin. TSP is also NP-hard since computing the optimal tour requires evaluating all possible permutations of the sites, which takes $O(N!)$ time. Fortunately, there exist efficient TSP solvers such as the Google OR-Tools\cite{ortools}, a premier optimization toolkit with many tools such as CP-SAT, MPSolver, and more, that can efficiently determine the optimal or near-optimal tour. Therefore, in the first step, we utilize a TSP solver to compute the shortest tour for the two vehicles, which is outlined in Algorithm~\ref{alg:routing}, line 1 with $\mathcal{X}$ denoting the locations of all sites and $\mathcal{P}$ denoting the generated shortest tour. Notably, the tour generated by the TSP solver provides overall guidance for planning paths for UGV and UAV. However, the actual paths of the two vehicles are generated by also incorporating other factors such as the UAV's energy allocation at each site, the UAV's take-off point, the two vehicles' rendezvous point, etc.  

\begin{algorithm}[h]
\SetAlgoLined

\tcp{Solve TSP to generate a tour} 
$\mathcal{P} = \texttt{TSP-solver}(\mathcal{X}$) \label{line:tsp}\\
\tcp{MCTS for energy allocation} 
{
  Create root node $v_0$\\ \label{line:rootnode}
  {
  \While{Maximum number of iterations not reached} 
  {
      \tcp{MCTS selection} 
      $v_i \leftarrow \texttt{UCB\_Selection}~(\text{Tree},v_0)$ \\
      \label{MCTS:line:Selection1}
      \eIf{$\operatorname{level}(v_i) <$ \emph{terminal}  {\bf and} $m(v_i) = 0$ }{
         \tcp{MCTS expansion}
         
         Tree $\leftarrow$ \texttt{Expand}~(\text{Tree}, $v_i$, \text{Energy})
         \label{MCTS:line:Expand1}
         
         \If{already expanded}{\textbf{continue}}
         \label{MCTS:line:Expand2}
      }{
         \tcp{MCTS rollout}
         $C \leftarrow \texttt{Rollout}~(v_i)$\;
         \label{MCTS:line:rollout}
      }
      \tcp{MCTS backpropagation}\
     \label{MCTS:line:Backpropagation1} {\While{$v_i \neq \mathrm{NULL}$}{
        \tcp{Update rewards and counts}
         $\alpha(v_i) \leftarrow \alpha(v_i) + \alpha$\\
         $n(v_i) \leftarrow n(v_i) + 1$\\ 
         $v_i\leftarrow$ parent of $v_i$
      }}\label{MCTS:line:Backpropagation2}
      $M \leftarrow M + 1$
}\label{line:endwhile}
}
\texttt{Path$\_$Generation} ($v,\mathcal{P}$) \label{MCTS:line:PathGeneration}
}
\caption{Energy-aware UGV-UAV routing}
\label{alg:routing} 
\end{algorithm}
\vspace{-5 mm}
\subsection{The second step of Algorithm~\ref{alg:routing}: MCTS}
\vspace{-1 mm}
The second step of Algorithm~\ref{alg:routing} focuses on solving the second subproblem that asks for computing the best energy allocation for the UGV across each site (Lines~\ref{line:rootnode} - \ref{line:endwhile}). 

Algorithm~\ref{alg:routing}'s first step generates a tour that specifies the order of surveying the sites. Since we consider the discrete energy levels for the UAV and the sites are surveyed in sequential order, the second subproblem can be treated as a tree search problem where we select an energy level for the UAV at each tree level. The depth of the tree depends on the number of sites. Clearly, this tree search problem is also NP-hard, which requires evaluation of all possible combinations of the UAV's energy levels along the TSP tour to find the optimal solution. One optimal solution could be the depth-first search (DFS) algorithm. However, the exponential growth of the tree makes the DFS algorithm not practical when the number of sites and the number of energy levels become large.


Instead, we utilize the Monte Carlo Tree Search (Algorithm~\ref{alg:routing}, lines~\ref{line:rootnode} - \ref{line:endwhile}). It has been shown that MCTS is an efficient algorithm that finds near-optimal solutions for tree search problems~\cite{kartal2016monte,browne2012survey}.  
To this end, we employ MCTS to find the best energy allocation at each site such that the total mission time is minimized and the energy constraints of the vehicles are satisfied. Next, we introduce the tree structure, pruning strategies, and the MCTS algorithm in detail. 
\vspace{-2 mm}
\subsubsection{Tree structure}
\vspace{-1 mm}
The tree starts at a \textit{root node}, which denotes the commencement location for the cooperative system of the UGV and UAV. The root node also stores the initial energy reserves for the two vehicles, setting the stage for subsequent energy-aware planning. 
The tree's \textit{child nodes} capture possible energy allocation options for the UAV at a particular site. 
Each child node represents a different energy allocation 
at a site. 
For instance, if the UAV can allocate $[20\%, 40\%, 60\%]$ of its remaining energy, three distinct child nodes (or branches) are spawned and each corresponds to one of the three energy percentages. Recall that each energy level corresponds to a range or a circle with the site as the center and the range as the radius (Section~\ref{subsec:coop_opera}). That way, each node also represents a certain range for a site.  As the UAV and UGV progress through sites, the cumulative influence of prior energy allocations becomes apparent in ensuing branches, progressively restricting future choices due to waning energy reserves.  In addition, every node stores data such as elapsed mission time, remaining energy for UGV and UAV, and the energy allocated.


\vspace{-2 mm}
\subsubsection{Tree pruning}
\vspace{-1 mm}
Tree pruning serves as a method to speed up tree search. 
It reduces computational demands and emphasizes paths that are both feasible and optimal. We introduce the two pruning strategies as follows. 
\vspace{-2 mm}
\paragraph{Optimality-based pruning}
Once a tree trajectory has been completely traversed, it denotes the attainment of a feasible, albeit not necessarily globally optimal plan. The total time taken is recorded. If, in subsequent explorations, the current node's cumulative time already surpasses this recorded minimum before the tree is fully expanded, then this node and its children are deemed sub-optimal and are consequently pruned from further consideration. This method aids in circumventing paths that are likely to be less efficient than those already discovered.
\vspace{-3 mm}
\paragraph{Constraint-based pruning}
During the tree expansion phase, if a node suggests that either the UAV or UGV has depleted its energy reserve, such a node, along with its subsequent child nodes, is pruned and not included in further exploration. This approach ensures that only scenarios where both vehicles retain operational energy are pursued. 



\subsubsection{MCTS}
\vspace{-1 mm}
Traversing the tree entails making iterative selections of possible energy allocations. This process begins by selecting a node and then expanding it. Subsequently, we proceed by randomly progressing through the respective child nodes until an end-point is reached. Once this end-point is achieved, both the reward and visit count are updated, tracing back to the tree's root. This traversal process can be categorized into four distinct stages: selection, expansion, rollout, and backpropagation. We will go into each of these stages in detail.

\paragraph{Selection (Algorithm~\ref{alg:routing}, line~\ref{MCTS:line:Selection1})}
In the section step, we use the Upper Confidence Bound (UCB) value as a guiding principle to choose an energy allocation for the UAV. It balances the tradeoff between the exploration of new nodes of energy allocation and the exploitation of already visited nodes. 
Particularly, in the UCB formula (\ref{UCB})
\begin{equation}
    \text{UCB}(v_i) =  \frac{\alpha(v')}{n(v')}+\texttt{const}\sqrt{\frac{2\ln N}{n(v')}},
    \label{UCB}
\end{equation}
    $\text{UCB}(v_i)$ is the Upper Confidence Bound for a particular energy allocation node \(v_i\), representing a certain UAV energy level allocated, \(\alpha(v')\) is the reward calculated from the total time for the child node \(v'\), \(n(v')\) denotes the number of times the child node \(v'\) has been visited, \(N\) denotes the cumulative number of times the parent node \(v_i\) has been visited, and \(\texttt{const}\) is a predefined constant that fine-tunes the balance between exploration of new allocations and exploitation of previously identified allocations.


\paragraph{Expansion (Algorithm~\ref{alg:routing}, lines~\ref{MCTS:line:Expand1}-~\ref{MCTS:line:Expand2})}
Upon selecting a node using the UCB formula, the next step is expansion. Specifically, the selected node is expanded based on the available energy allocation options. 
For instance, if three energy levels exist, and the remaining energy of both UAV and UGV permits, the chosen node will yield three children. Notably, this expansion for a particular node occurs only once.
\begin{figure}[h]
\centering{
\subfigure[Rollout ends early.]{\includegraphics[width=0.4\columnwidth]{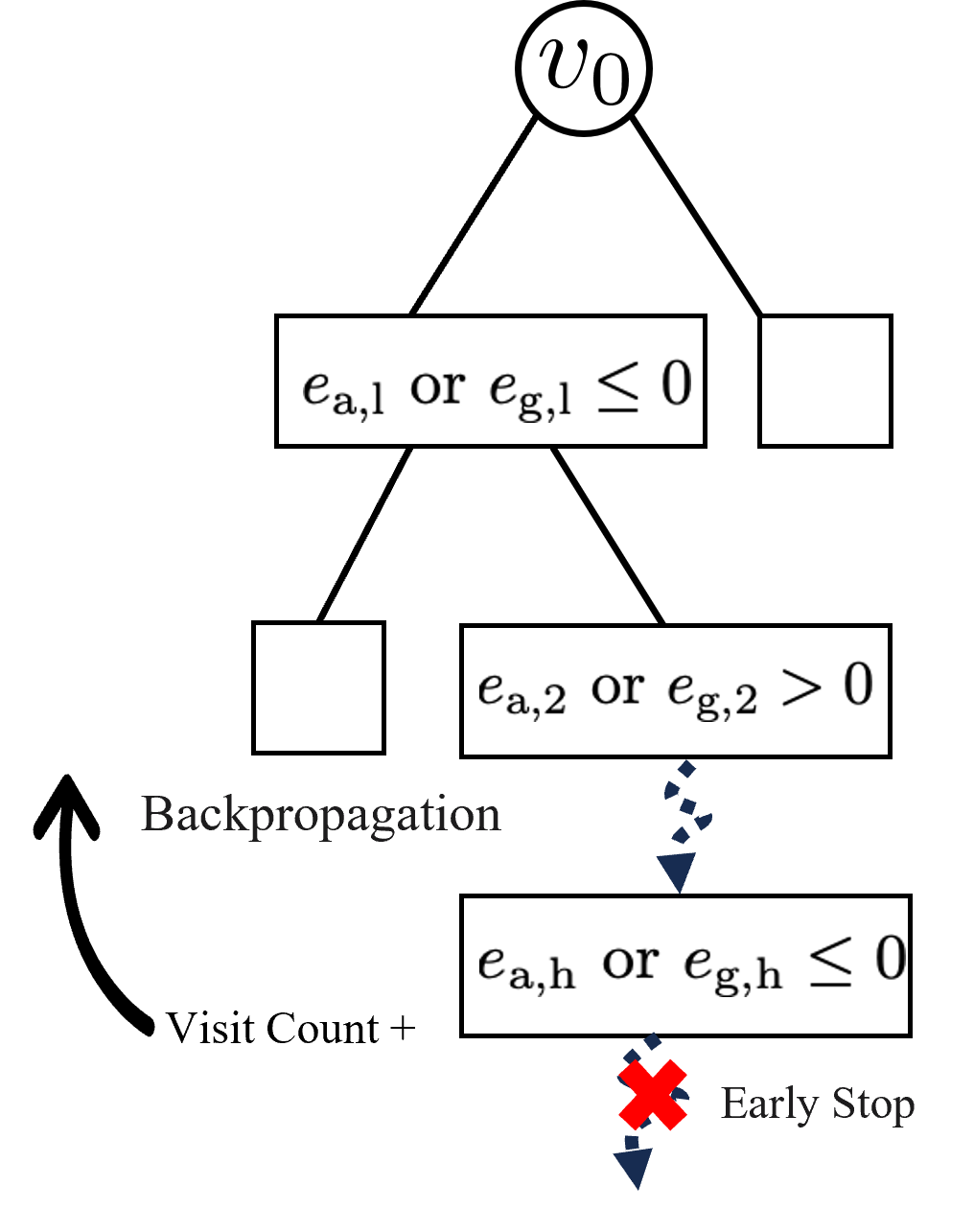}}
\subfigure[Rollout reaches the end.]{\includegraphics[width=0.4\columnwidth]{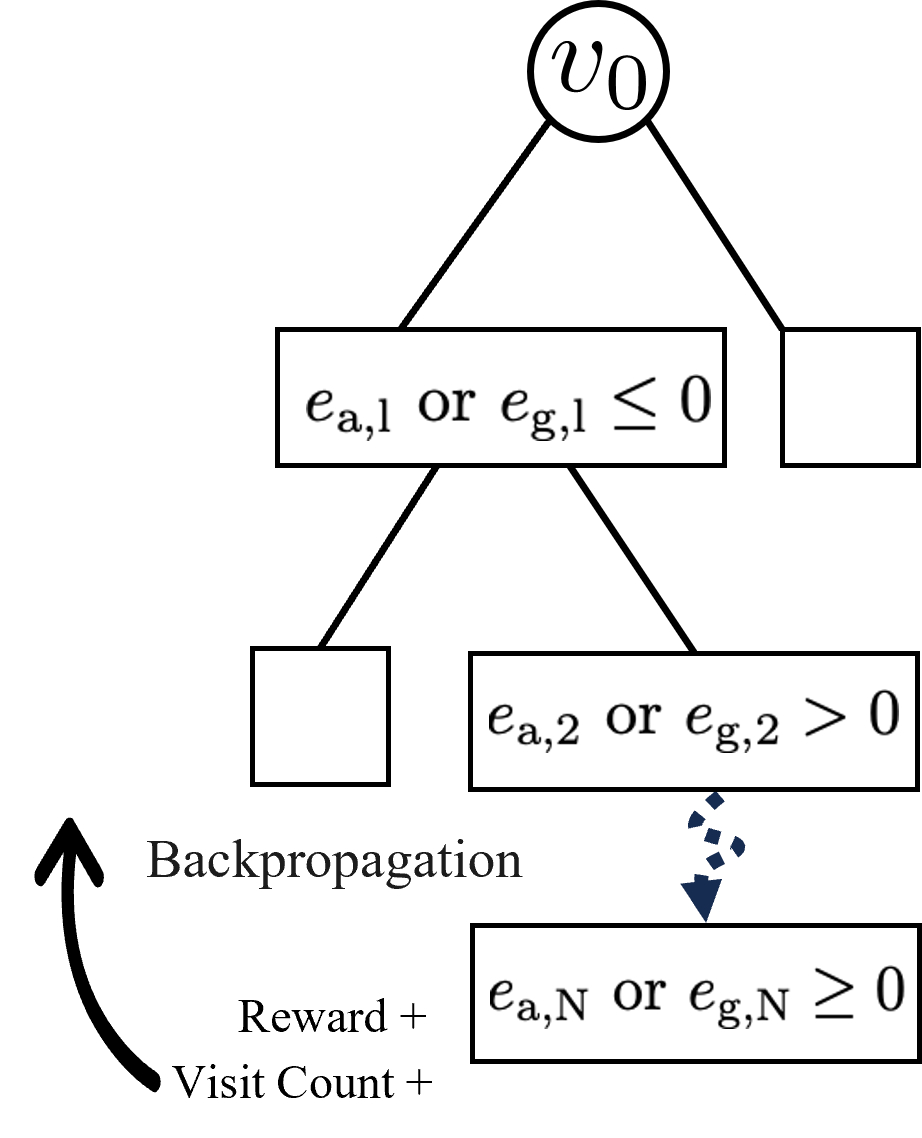}}
}
\caption{MCTS's rollout step. (a): A rollout ends early because either the UAV or UGV is not able to proceed due to low energy. We only perform the backpropagation of the visit count, i.e., increasing the visit counts of the nodes along the trajectory. (b): A rollout reaches the end stage and we perform a normal backpropagation of both the reward and visit count.}
\label{fig:rollout}  
\vspace{-5mm}
\end{figure}
\hspace{-14mm}
\paragraph{Rollout (Algorithm~\ref{alg:routing}, line~\ref{MCTS:line:rollout})}
After the expansion, the rollout phase simulates forward from the newly expanded node, selecting actions randomly until reaching the end stage or when either the UGV or UAV exceeds the energy limit. If the energy of the UGV at step \(h\), denoted as \(e_\text{g,h}\) or the energy of the UAV at step \(h\), denoted as \(e_\text{a,h}\), reaches zero before the end stage, we end the rollout early. In this case (Figure \ref{fig:rollout}-(a)), we backpropagate the visit count only (i.e., increasing the visit counts of the nodes along the trajectory), which discourages the future selection of the trajectory. In contrast, if a rollout is able to reach the end stage, we perform a normal backpropagation of both the reward and visit count. This rollout procedure optimizes for total mission time while ensuring energy constraints.

\paragraph{Backpropagation (Algorithm~\ref{alg:routing}, lines~\ref{MCTS:line:Backpropagation1}-\ref{MCTS:line:Backpropagation2})}

Upon reaching a terminal state, a reward is determined based on the total time spent. The reward, denoted as $\alpha$, is defined as the inverse of the total time multiplied by a tunable parameter $K>0$, that is, $\alpha = \frac{K}{T_\text{total}} $. Then the reward $\alpha$ is propagated back up the tree root, updating both the rewards and visit counts for the traversed nodes.

\vspace{-5 mm}
\subsection{Path generation from MCTS}
\begin{figure}
\centering
\subfigure{\includegraphics[width=0.35\columnwidth]{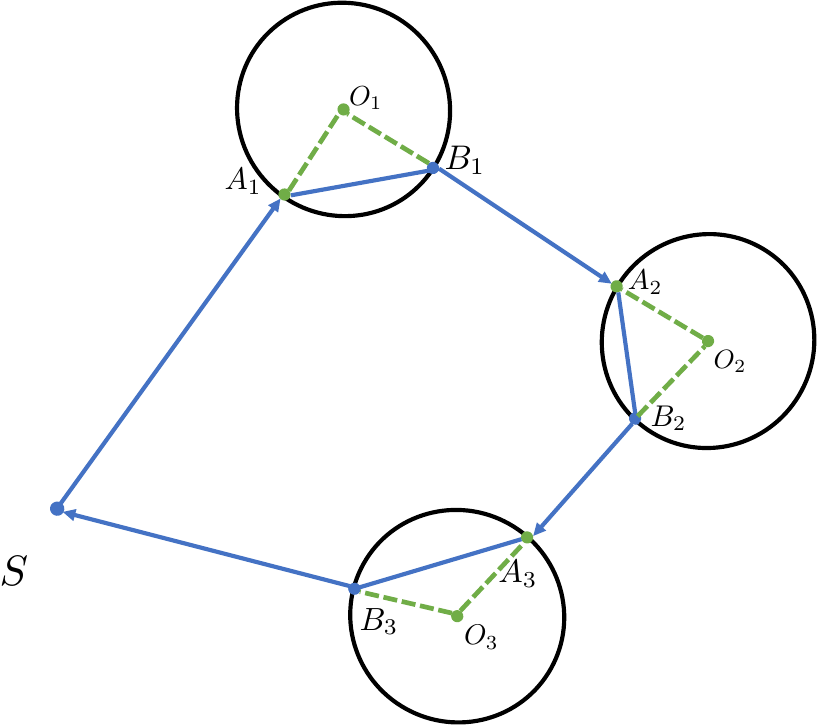}}
\caption{
An illustration of transforming the TSP tour \(S \rightarrow O_1 \rightarrow O_2 \rightarrow O_3 \rightarrow S\) into the UGV path \(S \rightarrow A_1 \rightarrow B_1 \rightarrow A_2 \rightarrow B_2 \rightarrow A_3 \rightarrow B_3 \rightarrow S\). 
}
\label{fig:4p} 
\end{figure}

\vspace{-1 mm}
The MCTS generates a combination of energy levels or ranges for the UAV across all sites. Combined with the generated TSP tour, these ranges can be decoded to paths for both the UAV and UGV (as outlined in Algorithm~\ref{alg:routing}, line~\ref{MCTS:line:PathGeneration}). 
By integrating this with the TSP tour, \(\mathcal{P}\), we can pinpoint the entry and exit locations of the UGV at each circle Figure~\ref{fig:Tri}, consequently shaping a chord path within the circle.
The TSP solver generates a survey tour for UGV, determining the sequence of sites to be visited. This sequence can then be converted into a feasible UGV path. As shown in Figure~\ref{fig:4p}, the UGV path is initialized by connecting all the sites using the TSP tour, i.e., \(S \rightarrow O_1 \rightarrow O_2 \rightarrow O_3 \rightarrow S\). Recall that each site corresponds to a circle centered at the site and with the radius converted from the energy allocated. 
The UGV path is then computed as the polyline from the start location to each intersection of the circles with the initialized path, and finally back to the start location, i.e., \(S \rightarrow A_1 \rightarrow B_1 \rightarrow A_2 \rightarrow B_2 \rightarrow A_3 \rightarrow B_3 \rightarrow S\).

\vspace{-5mm}

By factoring in their speed differentials and the survey time, we can calculate the UGV's position on the chord after the UAV's survey completion. Using all the information above, we can compute the meeting point of the two vehicles on the chord. Thus, the paths for both the UAV and UGV can be determined.

As shown in Figure~\ref{fig:Tri}, to calculate the rendezvous point \(P\) located on \(AB\), we set the unknowns to be \(|MP|\) and \(|OP|\). Given \(M\) is the midpoint of the chord \(AB\), the triangle \(OMP\) is right-angled at \(M\). Then we have
\begin{equation*}
|MP|^2 = |OP|^2 - |OM|^2,
\end{equation*}
and 
\begin{equation}
|MP| = \sqrt{|OP|^2 - |OM|^2}.
\label{eq:MP}
\end{equation}
As \(|AO|\) (the radius) and \(|AB|\) (from the TSP tour) are known, we compute \(|OM|\) by
\begin{equation}
|OM| = \sqrt{r^2 - \left( {|AB|}/{2} \right)^2}.
\label{eq:OM}
\end{equation}
Substituting Equation~\ref{eq:OM} into Equation~\ref{eq:MP}, we obtain
\begin{equation}
|MP| = \sqrt{|OP|^2 + \left( {|AB|}/{2} \right)^2 - r^2}.
\label{eq:MP-eq1}
\end{equation}
In addition, for the UGV and UAV to meet at point \(P\), the time that each vehicle spends to arrive at \(P\) from \(A\) must be equal. The UAV travels for \(\frac{r + |OP|}{V_\text{a}}\) time (with \(AO = r\)) along its path \(A \to O \to P\) and takes \(T_\text{survey}\) time for the survey at \(O\). In total, it spends \(\frac{r + |OP|}{V_\text{a}} + T_\text{survey}\) to reach \(P\). For the UGV, it will end on either \(AM\) or \(MB\). If it ends on \(AM\), the UGV spends \(\frac{AP}{V_\text{g}} = \frac{AM \pm |MP|}{V_\text{g}} = \frac{\frac{|AB|}{2} - |MP|}{V_\text{g}}\) to reach \(P\). Making these two times equal, we have:
\begin{equation}
\frac{r + |OP|}{V_\text{a}} + T_\text{survey} = \frac{\frac{|AB|}{2} - |MP|}{V_\text{g}}.
\label{eq:time-eq21}
\end{equation}

If it ends on \(MB\), the UGV spends \(\frac{AP}{V_\text{g}} = \frac{AM \pm |MP|}{V_\text{g}} = \frac{\frac{|AB|}{2} + |MP|}{V_\text{g}}\) to reach \(P\). Similarly, we have:
\begin{equation}
\frac{r + |OP|}{V_\text{a}} + T_\text{survey} = \frac{\frac{|AB|}{2} + |MP|}{V_\text{g}}.
\label{eq:time-eq22}
\end{equation}

Now, we can use Equations~\ref{eq:MP-eq1} and~\ref{eq:time-eq21} or Equations~\ref{eq:MP-eq1} and~\ref{eq:time-eq22} to solve for the two unknowns \(|MP|\) and \(|OP|\). However, there can only be one solution as the UAV's travel time must be the same as that of the UGV. Therefore, there must be unrealistic negative values from solving either Equations~\ref{eq:MP-eq1} and~\ref{eq:time-eq21} or Equations~\ref{eq:MP-eq1} and~\ref{eq:time-eq22}. Once we have either \(|MP|\) or \(|OP|\), we can calculate the coordinates of point \(P\) since the coordinates of \(A\), \(B\), \(M\), and \(O\) are known.

Notably, depending on the speeds of the UGV and UAV, there may be cases where the UGV must wait for the UAV at point \(B\), introducing a waiting time that must be incorporated into the total mission time, \( T_\text{total} \).

\begin{figure*}[t]
\centering{
\subfigure[Algorithm~\ref{alg:routing}, $N=5$, $T_\text{total}=19.27$ h,  $T_\text{run} = 0.18$ s.]{\includegraphics[width=0.32\columnwidth]{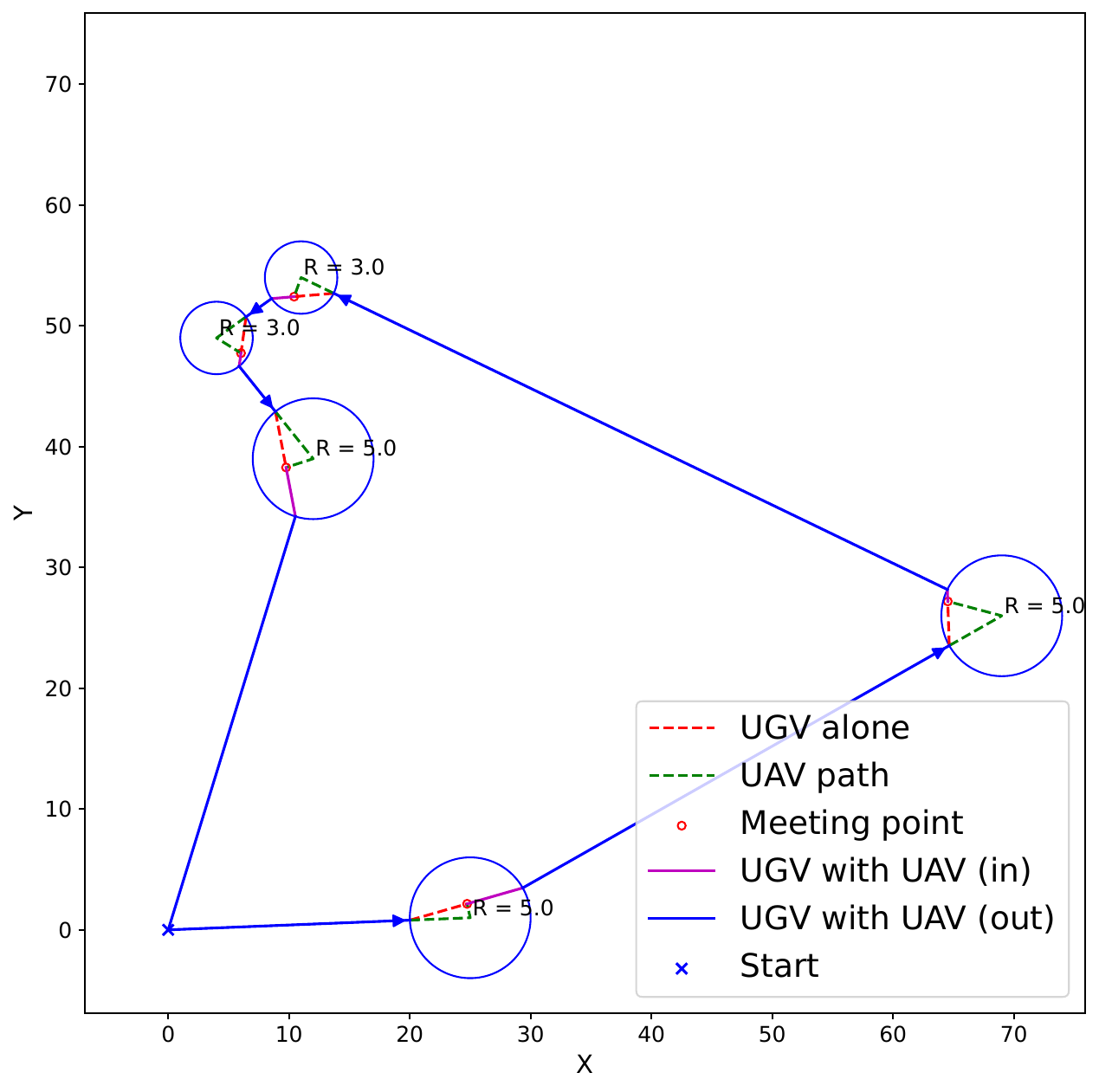}\hspace{1mm}}
\subfigure[Algorithm~\ref{alg:routing}, $N$ = 12, $T_\text{total} = 22.33$ h,$T_\text{run} = 0.52$ s.]{\includegraphics[width=0.32\columnwidth]{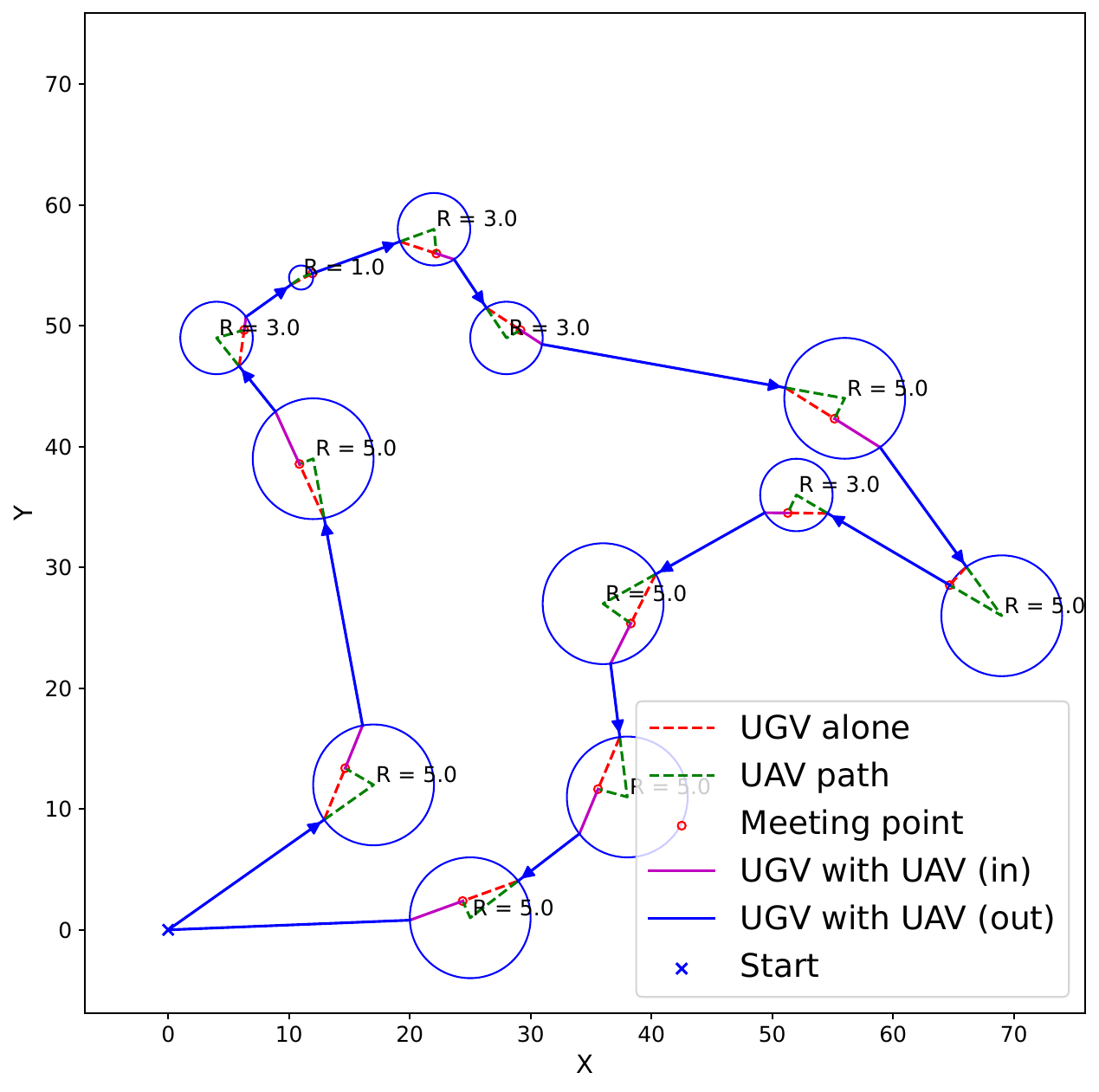}\hspace{1mm}}
\subfigure[Algorithm~\ref{alg:routing}, $N=14$ , $T_\text{total} = 37.21$ h, $T_\text{run} = 0.58$ s.]{\includegraphics[width=0.32\columnwidth]{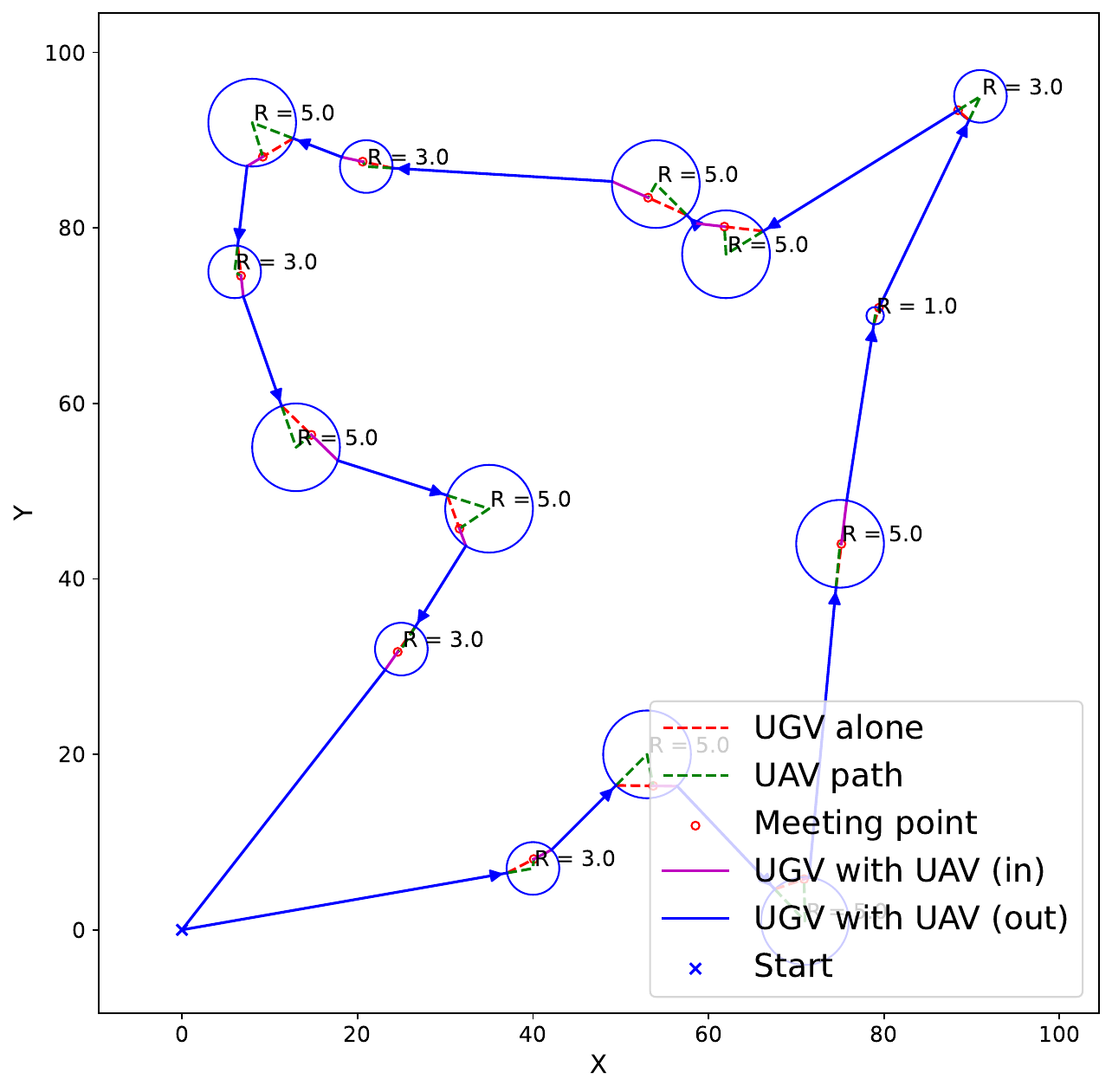}}
\subfigure[\texttt{Brute force}, $N=5$ , $T_\text{total}$=19.27 h, $T_\text{run} =  0.6$ s.]{\includegraphics[width=0.32\columnwidth]{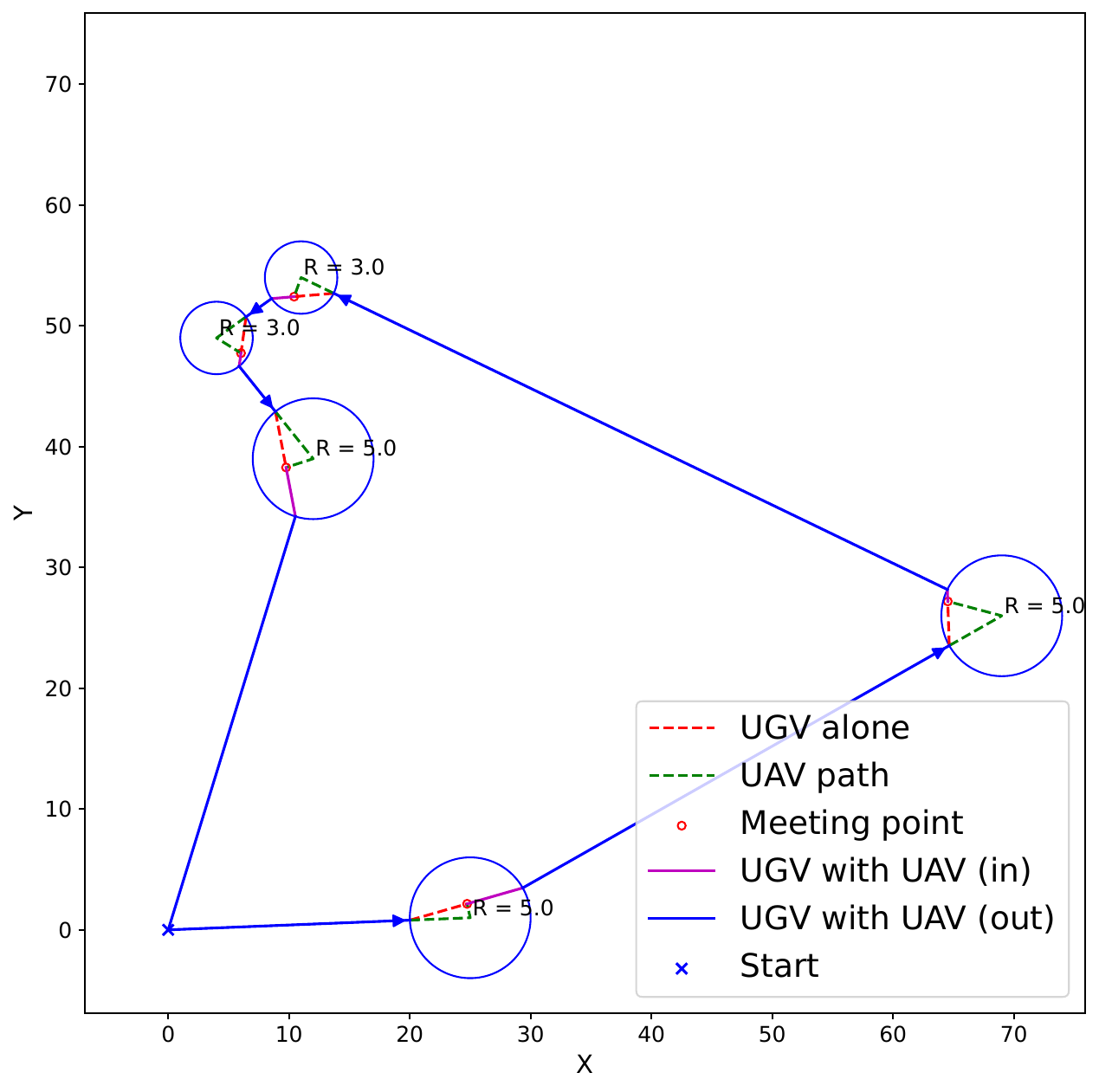}\hspace{1mm}}
\subfigure[\texttt{TSP+DFS}, $N=12$ , $T_\text{total}$ = 22.21 h, $T_\text{run} =  13.43$ s.]{\includegraphics[width=0.32\columnwidth]{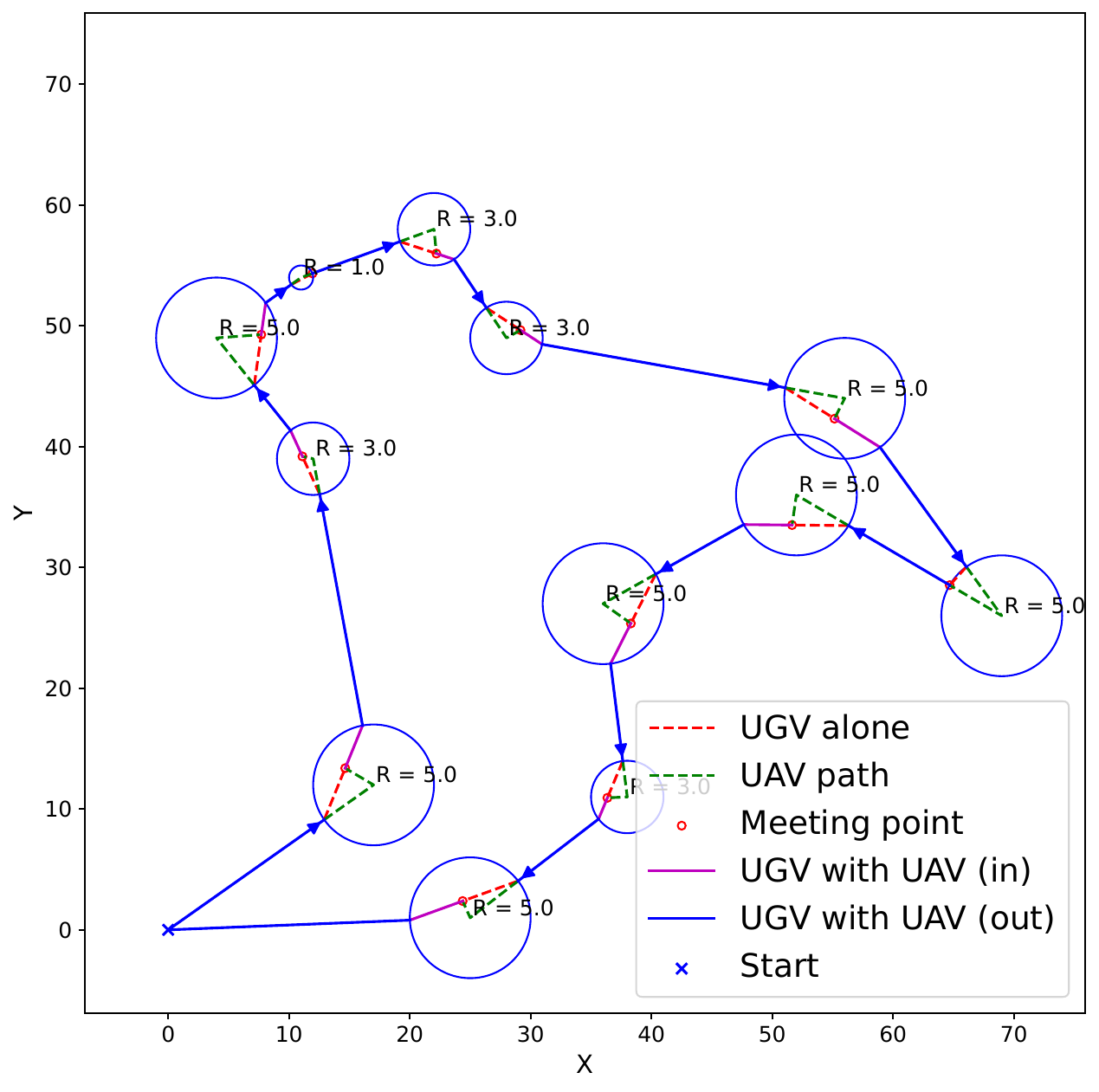}\hspace{1mm}}
\subfigure[\texttt{TSP+DFS}, $N=14$ , $T_\text{total}$ = 37.19 h, $T_\text{run} =  162.45$ s.]{\includegraphics[width=0.32\columnwidth]{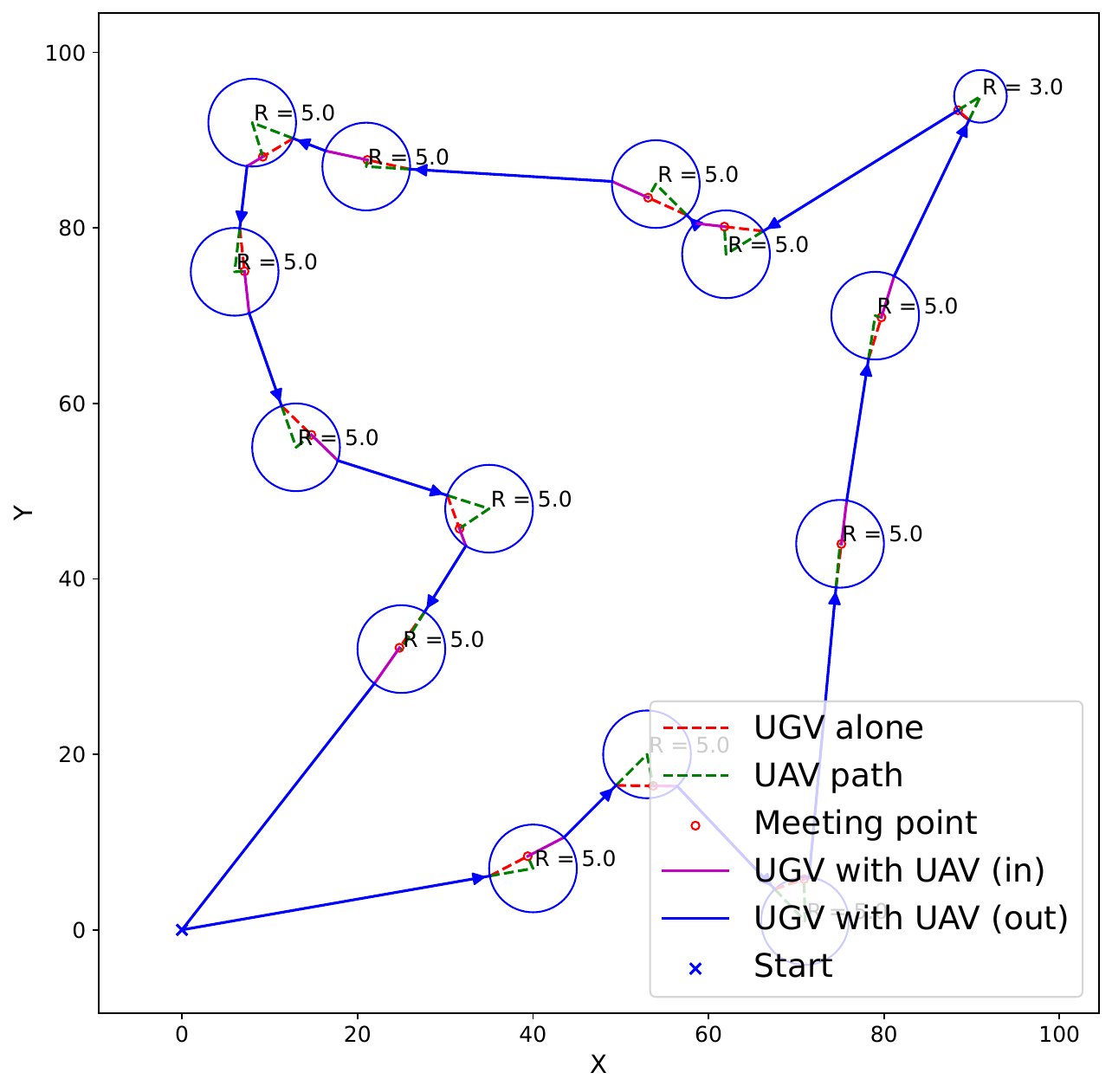}}
}
\caption{
Qualitative comparison of the tour generated by Algorithm~\ref{alg:routing} with those generated by \texttt{Brute force} and \texttt{TSP+DFS} across $N=5, 12, 14$ sites.}
\label{fig:Qualitative_result}
\vspace{-2mm}
\end{figure*}

\vspace{-5mm}
\section{Results}
This section presents both qualitative and quantitative results to demonstrate the effectiveness and efficiency of our energy-aware routing algorithm (Algorithm~\ref{alg:routing}). Specifically, we compare Algorithm~\ref{alg:routing} to three other baseline algorithms. The first algorithm is a \texttt{brute force} algorithm that exhaustively enumerates all possible tours of surveying sites and all possible combinations of the UAV's energy allocations across sites to find the optimal solution. The second algorithm integrates the TSP solution (Algorithm~\ref{alg:routing}'s first step) to find a shortest tour and the DFS algorithm to find the best combination of the UAV's energy allocations. We name it \texttt{TSP + DFS} algorithm. Notably, both the \texttt{brute force} and \texttt{TSP + DFS} algorithms run in exponential time, thus not applicable for large-scale instances. To evaluate the performance of Algorithm~\ref{alg:routing} with a large number of sites and energy levels, we compare it to the third algorithm, named \texttt{naive} algorithm. In the \texttt{naive} algorithm, the UGV follows the generated TSP tour~\ref{line:tsp}, transporting the UAV to each site and waiting on-site while the UAV conducts the survey.

Notably, we consider only five discrete energy levels for the ease of simulations. Increasing the granularity of energy levels would introduce more variations in feasible tour configurations and corresponding energy allocations, potentially leading to finer optimizations but also increasing computational complexity.

All evaluations are performed on a Linux desktop with Ubuntu 20.04 powered by an AMD Ryzen 5600X processor with 32GB RAM.
\vspace{-5 mm}
\subsection{Qualitative results}
\vspace{-1 mm}
We first present qualitative results to evaluate the performance of Algorithm~\ref{alg:routing} across three different site numbers, i.e., \( N = 5, 12, 14 \) in Figure~\ref{fig:Qualitative_result}. Figure~\ref{fig:Qualitative_result}-(a), (b), (c) shows that Algorithm~\ref{alg:routing} effectively plans paths for the UGV and UAV by first generating a TSP tour and then allocating energy levels (or ranges) at each site using MCTS. Comparing to Figure~\ref{fig:Qualitative_result}-(d), (e), (f), we observe that the paths generated by Algorithm~\ref{alg:routing} are close to those generated by the \texttt{brute force} algorithm with $N=5$ (Figure~\ref{fig:Qualitative_result}-(d)), \texttt{TSP + DFS} algorithm with $N=12$ (Figure~\ref{fig:Qualitative_result}-(e)), and \texttt{TSP + DFS} algorithm with ($N=14$ Figure~\ref{fig:Qualitative_result}-(f)). In addition, Algorithm~\ref{alg:routing} performs on par with \texttt{brute force} and \texttt{TSP + DFS} in terms of the mission time $T_\text{total}$, but runs much faster. Especially, when $N=14$, the runtime $T_\text{run}$ of Algorithm~\ref{alg:routing} (0.58 s) is more than 280 times less than that of \texttt{TSP + DFS} (162.45 s).

\begin{figure}
\centering
\subfigure{\includegraphics[width=0.5
\columnwidth]{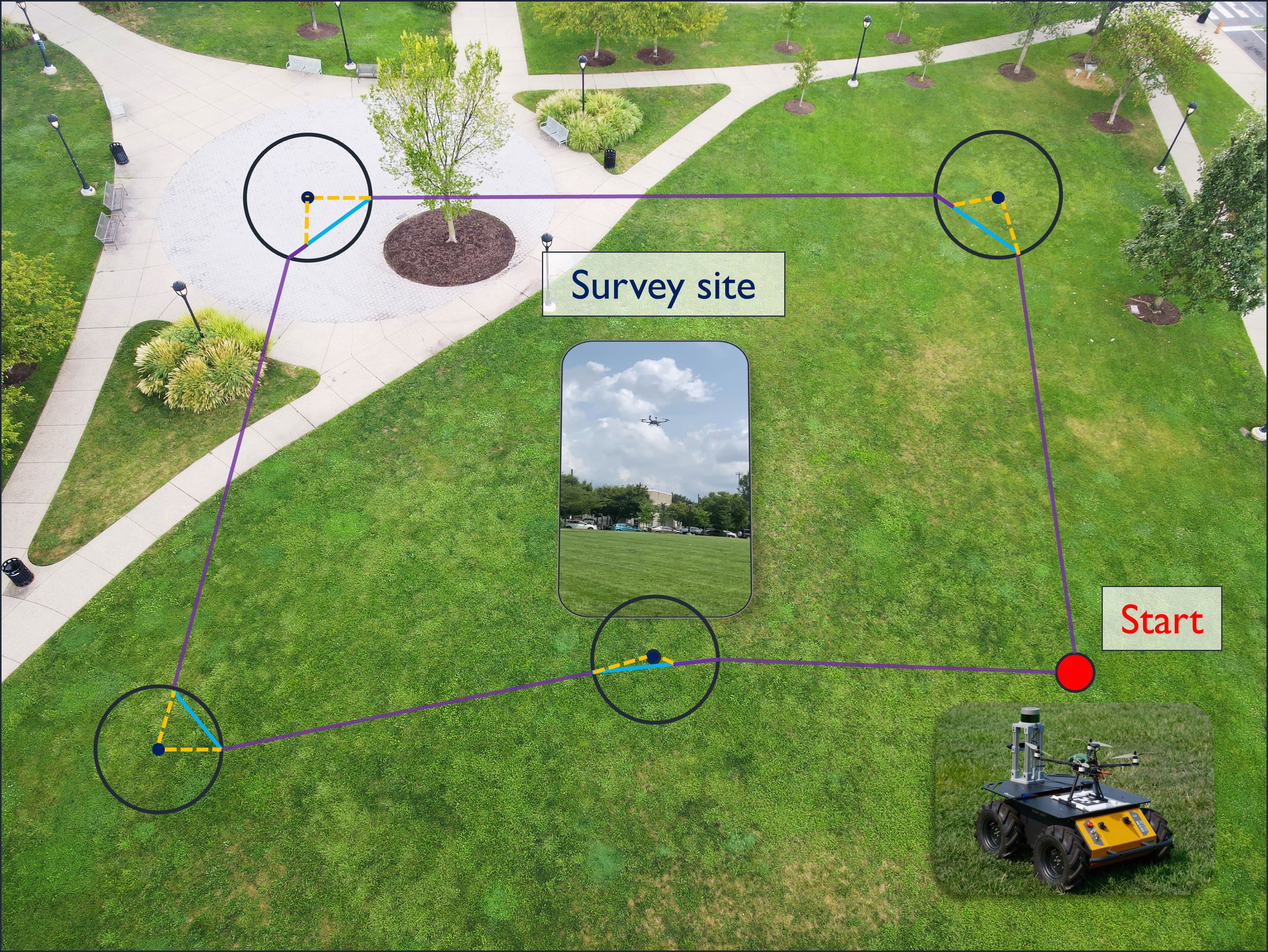}}
\caption{An illustration of the proof-of-concept field experiment. The solid purple lines represent the paths where the UGV ferries the UAV. The cyan solid lines represent the paths of the UGV without the UAV on it. The dashed yellow lines represent the flying paths of the UAV.
}
\label{fig:Finished} 
\vspace{-5mm}
\end{figure}

\textit{Real robot experiment:} We also run a proof‐of‐concept field experiment to demonstrate the effectiveness of Algorithm~\ref{alg:routing} for computing tours for the two vehicles. Our cooperative autonomous system comprises a modified Clearpath husky rover and a ModalAI Sentinel drone. The mission of the two vehicles is to survey five sites in Drexel Park (Figure~\ref{fig:Finished}).
\vspace{-5mm}
\subsection{Quantitative results}
\begin{figure}[H]
\centering{
\subfigure{\includegraphics[width=0.49\columnwidth]{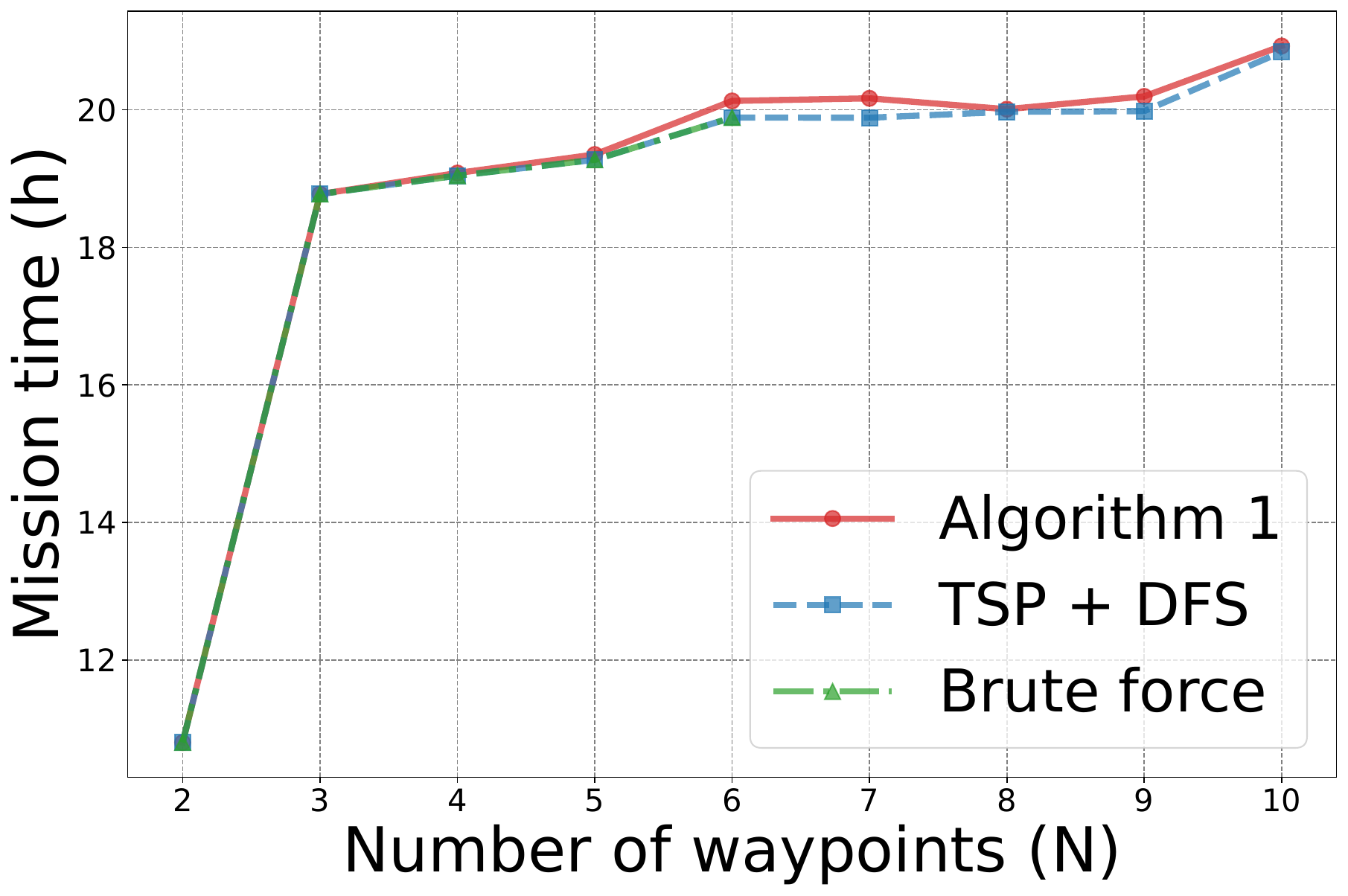}}
\subfigure{\includegraphics[width=0.49\columnwidth]{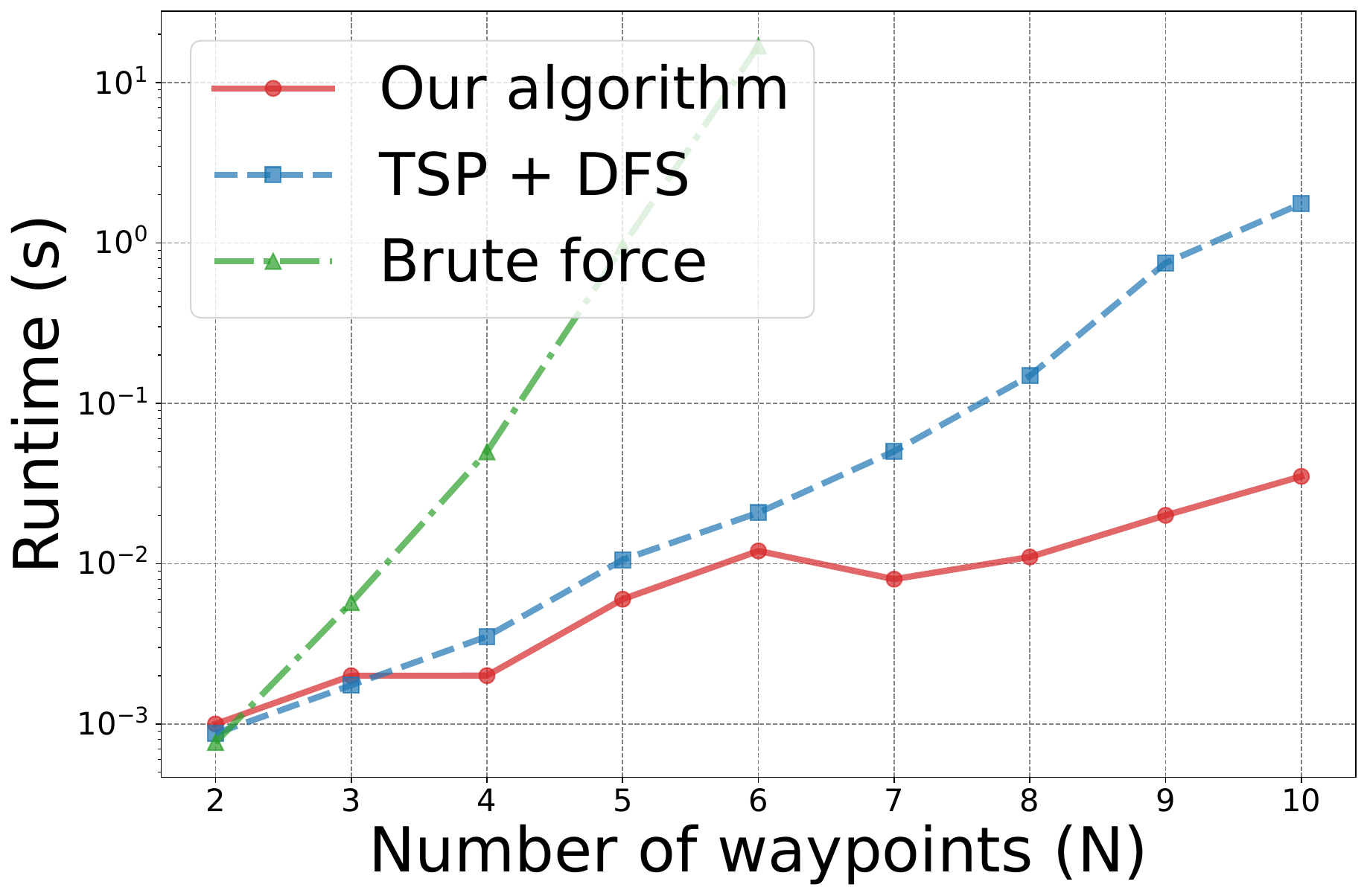}}
}
\caption{
 Quantitative composition of Algorithm~\ref{alg:routing} with \texttt{brute force} and \texttt{TSP + DFS} in terms of mission time $T_\text{total}$ and runtime $T_\text{run}$ in small scale cases. The y-axis is in \texttt{log} scale. \texttt{brute force} stops at six sites due to its prohibitively long runtime when $N>6$. 
}
\label{fig:compare_small}  
\end{figure}

\begin{figure}[H]
\centering{
\subfigure{\includegraphics[width=0.49\columnwidth]{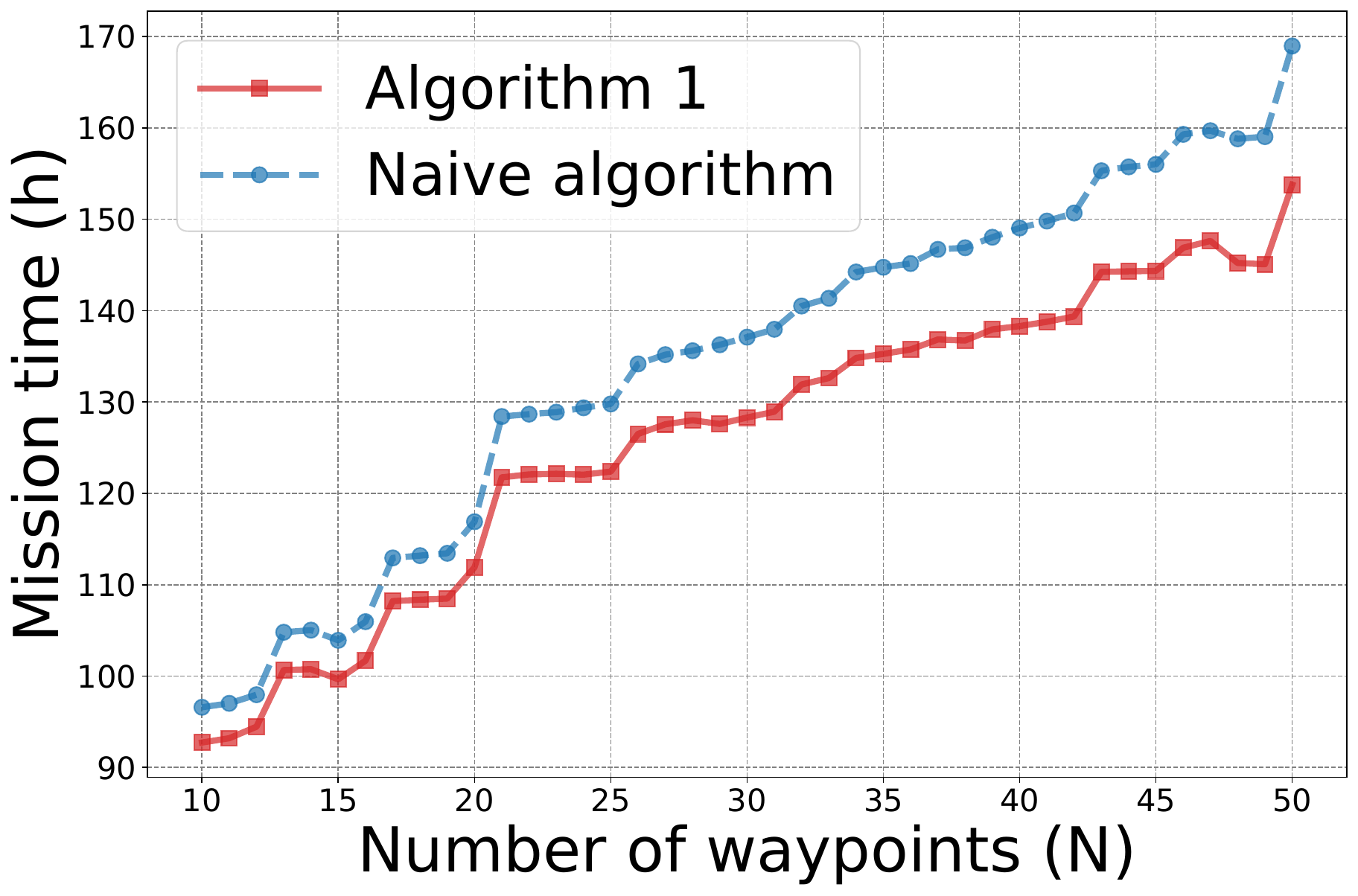}}
\subfigure{\includegraphics[width=0.49\columnwidth]{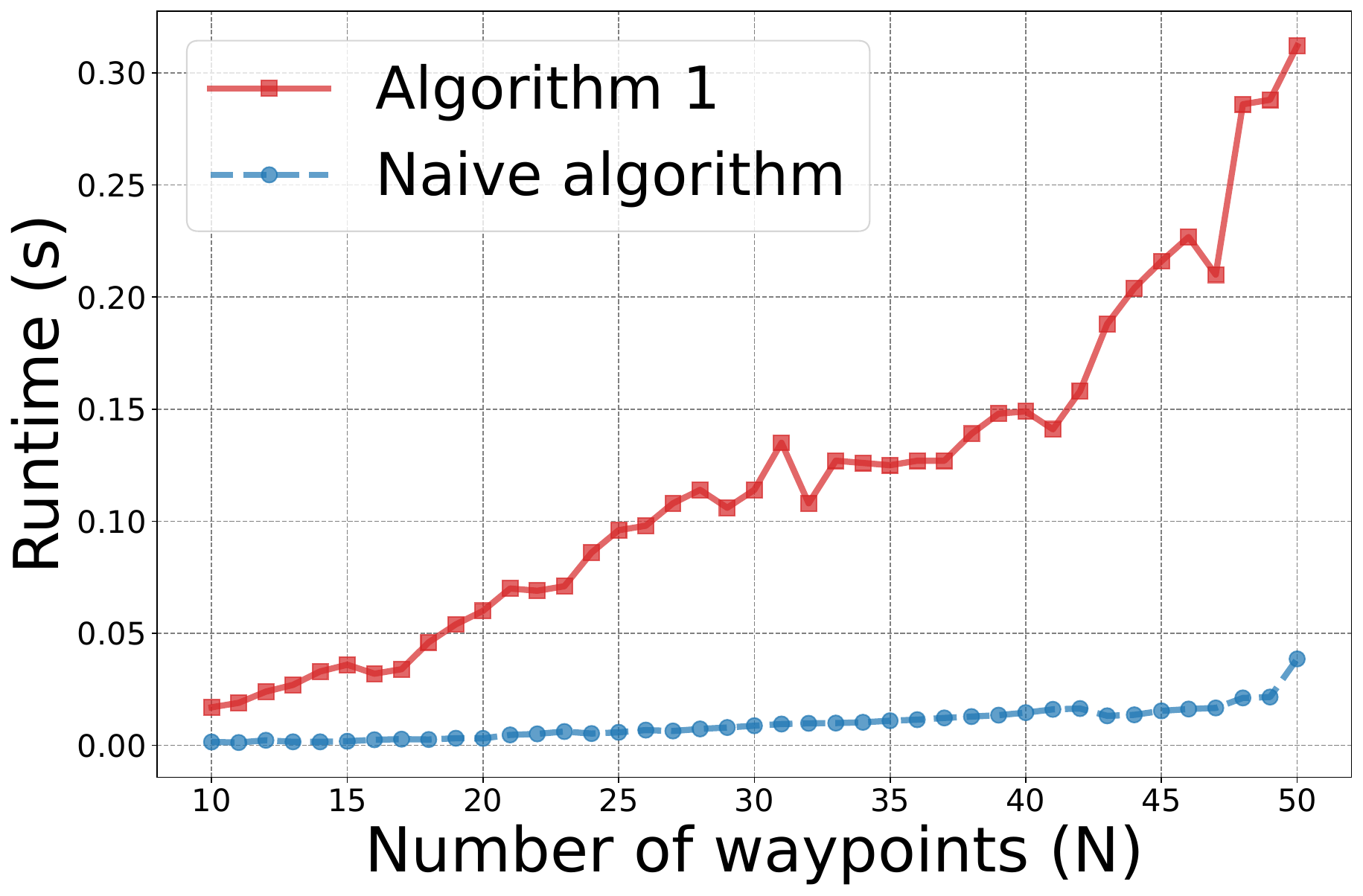}}
}
\caption{
Quantitative composition of Algorithm~\ref{alg:routing} with \texttt{naive} and in terms of mission time $T_\text{total}$ and runtime $T_\text{run}$ in large scale cases. The y-axis is in \texttt{log} scale.
}
\label{fig:compare_large}  
\end{figure}

We further evaluate Algorithm~\ref{alg:routing} through quantitative comparisons with other baseline algorithms. Particularly, in small-scale instances with a small number of sites $N\in[2,\cdots, 10]$, we compare Algorithm~\ref{alg:routing} to \texttt{brute force} and \texttt{TSP + DFS}, shown in Figure~\ref{fig:compare_small}. When the number of sites increases to a large number, \texttt{brute force} and \texttt{TSP + DFS} become inapplicable. Then, we compare Algorithm~\ref{alg:routing} to the \texttt{naive} algorithm with $N\in[10,\cdots, 50]$, as shown in Figure~\ref{fig:compare_large}. 


The small-scale comparison, as shown in Figure~\ref{fig:compare_small} further demonstrates that Algorithm~\ref{alg:routing} achieves almost the same mission time but runs much faster than both \texttt{brute force} and \texttt{TSP + DFS}, especially when $N>5$. It is also observed that \texttt{TSP + DFS} performs close to \texttt{brute force}, which justifies the use of the TSP solver to first compute a guided tour for the UGV and UAV.

In the large-scale comparison, Algorithm~\ref{alg:routing} consistently outperforms the \texttt{naive} algorithm. Particularly, when the number of sites $N=50$, Algorithm~\ref{alg:routing} achieves a mission time that is 15.18 h less that of \texttt{naive}. In addition, even with $N=50$, Algorithm~\ref{alg:routing} runs less than 0.325 s to generate paths for UGV and UAV.

    

\vspace{-6mm}

\vspace{-10mm}
\section{Conclusion and Future Directions}

In our exploration of UAV-UGV cooperative missions, we introduced a planning methodology attuned to the energy budgets of both aerial and ground units. In scenarios where the UGV serves as both a transporter and a charging station for the UAV, our algorithm incorporates TSP to determine an initial tour and employs MCTS to refine the routes for both UGV and UAV. We evaluated the performance of our algorithm through simulations and experiments. Our findings demonstrated the algorithm's efficiency across a diverse range of instance sizes and its consistent production of near-optimal solutions.

An ongoing work is to refine our algorithm by merging the survey of nearby sites. This might change the structure of our decision tree but could provide performance gains on the total mission time. The second future work is to design an online routing algorithm based on our current offline algorithm, to offer real-time adaptability, catering to dynamically changing and uncertain environments. 
%
%
%
\vspace{-4mm}
\bibliographystyle{plain}
\bibliography{references}

\end{document}